\documentclass[acmtog,nonacm]{acmart}

\renewcommand\footnotetextcopyrightpermission[1]{}

\usepackage{multirow}
\usepackage[ruled]{algorithm2e}
\usepackage[capitalize]{cleveref}
\usepackage{amsmath}
\usepackage{bbm}
\usepackage{cancel}
\usepackage{afterpage}
\usepackage{makecell}

\SetAlFnt{\small}
\SetAlCapFnt{\small}
\SetAlCapNameFnt{\small}
\SetAlCapHSkip{0pt}

\crefname{section}{Sec.}{Secs.}
\Crefname{section}{Section}{Sections}
\crefname{table}{Tab.}{Tabs.}
\Crefname{table}{Table}{Tables}

\begin{document}

\title[Stylized Text-to-Motion Generation via Hypernetwork-Driven Low-Rank Adaptation]%
{Stylized Text-to-Motion Generation
via Hypernetwork-Driven Low-Rank Adaptation}

\author{Junhyuk Jeon}
\authornote{Equal contribution.}
\affiliation{
  \institution{Visual Media Lab, KAIST}
  \city{Daejeon}
  \country{Republic of Korea}
}
\email{jeonjh@kaist.ac.kr}

\author{Seokhyeon Hong}
\authornotemark[1]
\affiliation{
  \institution{Visual Media Lab, KAIST}
  \city{Daejeon}
  \country{Republic of Korea}
}
\email{ghd3079@kaist.ac.kr}

\author{Junyong Noh}
\authornote{Corresponding author.}
\affiliation{
  \institution{Visual Media Lab, KAIST}
  \city{Daejeon}
  \country{Republic of Korea}
}
\email{junyongnoh@kaist.ac.kr}

\begin{abstract}
Text-driven motion diffusion models are capable of generating realistic human motions, but text alone often struggles to express fine-level nuances of motion, commonly referred to as style. Recent approaches have tackled this challenge by attaching a style injection mechanism to a pretrained text-driven diffusion model. Existing stylization methods, however, either require style-specific fine-tuning of existing models or rely on heavy ControlNet-based architectures, limiting efficiency and generalization to unseen styles. We propose a lightweight style conditioning framework that dynamically modulates a pretrained diffusion model through hypernetwork-generated LoRA parameters. A style reference motion is encoded into a global style embedding, which is mapped by a hypernetwork to low-rank updates applied at each denoising step of the diffusion model. By structuring the style latent space with a supervised contrastive loss, our framework reliably captures diverse stylistic attributes, improves generalization to unseen styles, and supports optimization-based guidance without requiring predefined style categories. Experiments on the HumanML3D and 100STYLE datasets show state-of-the-art stylization results, while achieving improved stylization for unseen styles.
\end{abstract}



\begin{teaserfigure}
  \centering
  \includegraphics[width=\textwidth]{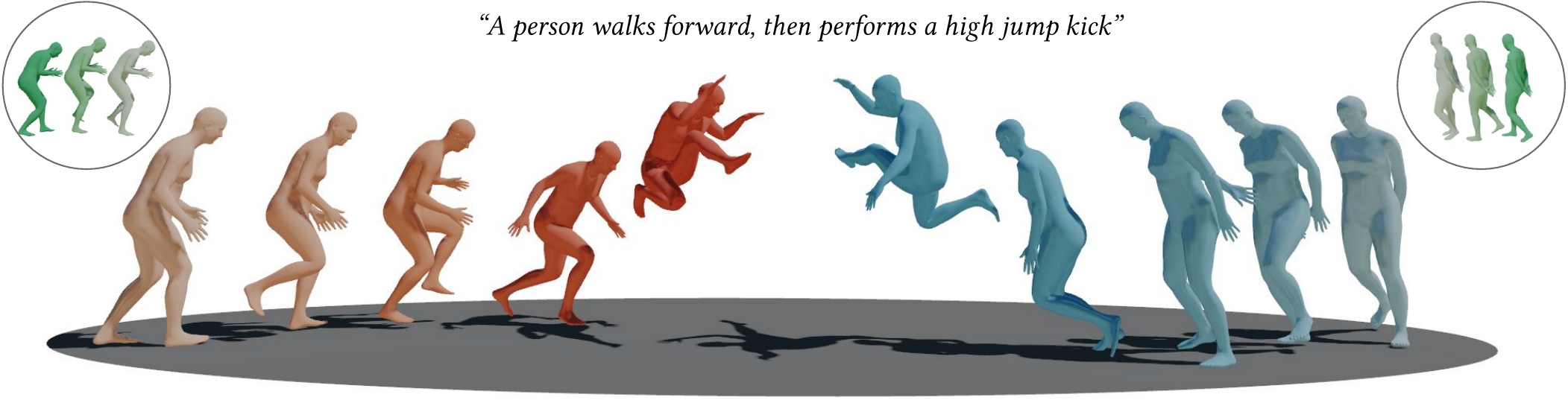}
  \caption{Our method enables flexible and efficient stylized text-to-motion generation with generalization to unseen motion styles.}
\end{teaserfigure}

\maketitle

\section{Introduction}
\label{sec:intro}
Advancements in text-driven human motion diffusion models have enabled the generation of high-quality motion from intuitive natural language descriptions~\citep{tevet2022human, zhang2024motiondiffuse}. Despite this achievement, text alone often fails to convey \textit{style} --- subtle attributes such as emotion or personality that shape how a specific \textit{action} is performed. Recent methods address this limitation by augmenting text-to-motion diffusion models with a style conditioning mechanism that extracts and injects style from a reference motion~\citep{zhong2024smoodi, sawdayee2025dance, wu2025semantically}. In this paradigm, text serves as the source of motion \textit{content}, while the reference motion serves as the source of motion \textit{style}. This approach enables the generation of nuanced and expressive motion that pure text-driven methods struggle to achieve.

Given the abstract nature of style, its learned representations vary across existing methods. Label-supervised approaches~\citep{zhong2024smoodi, sawdayee2025dance} use discrete style annotations, relying on datasets such as 100STYLE~\citep{mason2022real} to define style categories. While effective for styles seen during training, such methods introduce biases toward observed style labels, limiting generalization to unseen styles. In contrast, \citeN{wu2025semantically} forgoes explicit style labels and instead learns continuous style representations from unlabeled motion data~\citep{guo2022generating}. Although this approach reduces label-induced biases, the lack of explicit structural representation for motion style still limits generalization to styles beyond the training distribution.

Limitations in current stylized text-to-motion diffusion models also stem from their architectures. SMooDi~\citep{zhong2024smoodi} employs a ControlNet-based architecture~\citep{zhang2023adding} to inject style at each denoising step. While this design enables a single model to represent a wide range of styles, it substantially increases model capacity and computational overhead. Conversely, LoRA-MDM~\citep{sawdayee2025dance} adopts a parameter-efficient strategy by fine-tuning a pretrained diffusion model with Low-Rank Adaptation (LoRA) layers \citep{hu2022lora}. This approach is efficient in both training and inference, but its restriction to a single style per fine-tuning instance sacrifices scalability and generalization to diverse stylistic variations.

To address these challenges, we propose a novel hypernetwork-driven~\citep{ha2016hypernetworks} style conditioning framework that retains LoRA’s efficiency while enabling generalization to unseen styles, which we call \textit{HyperLoRA}. Given an arbitrary reference motion, a content-agnostic style adapter extracts a global style embedding and predicts a style-dependent LoRA update. These dynamically generated LoRA parameters are injected into a pretrained text-to-motion diffusion model at each denoising step, allowing expressive feature modulation with minimal computational and memory overhead.

To introduce semantic structure to the latent style space, we train the style adapter with supervised contrastive learning \citep{khosla2020supervised}, encouraging a discriminative, continuous, and generalizable style representation. This approach enables robust style extraction even from reference motions with unseen styles. Furthermore, we leverage the style adapter during inference to guide the optimization of intermediate diffusion latent variables. In contrast to classifier-guidance~\citep{dhariwal2021diffusion}, which relies on style classifiers trained on pre-defined discrete style categories, our approach enhances controllability by providing flexible style guidance. Extensive experiments demonstrate that our method achieves state-of-the-art stylized motion generation across standard benchmarks and diverse style references. Code is available at \url{https://github.com/junhyukjeon/style-salad}.

In summary, our contributions are as follows:
\begin{itemize}
    \item A structured latent style representation that enables generalizable style extraction from arbitrary reference motions, including unseen styles.
    \item A hypernetwork-driven LoRA conditioning mechanism for scalable and generalizable motion style modulation.
    \item A style-guided sampling strategy that extends traditional classifier guidance to continuous, instance-specific style embeddings extracted from reference motion.
\end{itemize}

\section{Related Work}
\label{sec:related}

\subsection{Text-Driven Human Motion Generation}
Text-driven human motion generation has emerged as a promising research area for its ability to synthesize complex human movements from intuitive natural language descriptions. Early works demonstrated the feasibility of this task by mapping text embeddings to pose sequences through sequence-to-sequence modeling or variational formulations~\citep{petrovich2022temos, guo2022generating, athanasiou2022teach}. While these approaches established a foundational connection between language and motion, they often suffered from limited motion diversity, temporal discontinuities, and a tendency toward over-smoothed or repetitive motion patterns.

Recent advances address these limitations through denoising diffusion models~\citep{ho2020denoising, song2020denoising}, which formulate motion generation as an iterative refinement process from a Gaussian noise conditioned on text \citep{tevet2022human, zhang2024motiondiffuse, kim2023flame}. By progressively denoising from stochastic latent representations, diffusion-based methods improve temporal coherence and sample diversity, producing motions that align more faithfully to textual semantics. Latent diffusion variants further improve computational efficiency by operating in low-dimensional learned motion embedding spaces rather than raw pose sequences \citep{chen2023executing, dai2024motionlcm, hong2025salad, sampieri2024length}.
In this work, we extend a pretrained text-driven motion diffusion model to reflect styles observed in a reference motion, enabling generation of stylized motions while faithfully following given textual descriptions.

\subsection{Motion Stylization}
Motion stylization refers to the process of modifying how a motion is performed, capturing stylistic attributes such as emotion and personality while preserving motion content. Historically, motion stylization has been studied primarily through motion style transfer, where style is transferred from a reference motion to a target sequence. Early works approached this task using example-based or optimization-driven formulations, often relying on paired or temporally aligned motion data to learn explicit mappings between content and style \citep{hsu2005style, min2010synthesis}. 

With advances in deep learning, subsequent works sought to reduce the reliance on carefully curated motion pairs by learning disentangled representations of motion content and style. Neural motion style transfer frameworks introduced feature-based losses and normalization mechanisms that inject style information from a reference motion into a target sequence without requiring explicit temporal alignment \citep{holden2016deep}. Follow-up work further improved runtime efficiency by replacing per-instance optimization with feed-forward networks \citep{holden2017fast}. More recent approaches extend this formulation by explicitly encoding content and style in separate latent representations, enabling unpaired training and flexible style transfer across unseen action-style combinations~\cite{aberman2020unpaired} and body parts~\cite{jang2022motion}. Related works further explore stylization in latent or canonical motion spaces, improving generalization and structural consistency~\cite{guo2024generative, zhang2024generative}, while zero-shot motion transfer methods remove dependency on paired data or explicit temporal alignment~\citep{raab2024monkey}.

Motion stylization has more recently expanded to generative settings, where style is injected directly into the motion synthesis process rather than applied as a post-processing step. In particular, recent studies began to adopt a pretrained motion diffusion model as a general-purpose backbone and fine-tune it to enable stylized motion generation. SMooDi \citep{zhong2024smoodi} introduces a ControlNet-based~\cite{zhang2023adding} conditioning mechanism that guides the denoising process using reference style motions. LoRA-MDM~\citep{sawdayee2025dance}, a concurrent work, adopts a parameter-efficient alternative by fine-tuning a pretrained motion diffusion model with low-rank adaptations~\cite{hu2022lora}, embedding style directly into the model parameters. Beyond these approaches, prior work has explored unsupervised style modeling~\citep{wu2025semantically} as well as multimodal stylization that incorporates multiple input modalities for richer control~\citep{guo2025stylemotif, zhong2025smoogpt}. In this work, we introduce a hypernetwork-driven~\cite{ha2016hypernetworks} LoRA that generalizes across diverse styles, with only a minimal increase in computational cost for style extraction from a reference motion.
\section{Preliminary}
\label{sec:preliminary}

\subsection{SALAD}
\label{sec:salad}
As a backbone of generative motion diffusion model, we employ SALAD~\citep{hong2025salad}, which is a skeleton-aware latent motion diffusion framework for text-conditioned human motion generation and editing. It encodes human motion sequences into structured latent representations using a skeleto-temporal variational autoencoder~(ST-VAE) that separates temporal and joint dimensions while modeling information exchange within each modality. Based on this representation, a denoiser is trained to generate text-conditioned latent features, which are subsequently decoded into motion sequences. By explicitly modeling multi-modal relationships through skeletal, temporal, and text-motion cross-attention layers, SALAD achieves state-of-the-art performance in text-driven human motion generation. These properties make it well suited for downstream adaptation tasks, such as stylized text-to-motion generation, where preserving semantic motion content is essential.

SALAD adopts the velocity parameterization~\citep{salimans2022progressive}, in which the denoiser predicts a velocity vector representing a linear combination of the clean latent and noise. Given a latent $z_0$ and noise $\epsilon \sim \mathcal{N}(0,I)$, the forward diffusion process is defined as
\begin{equation}
z_t = \alpha_t z_0 + \sigma_t \epsilon,
\end{equation}
and the denoiser is trained to predict the corresponding velocity
\begin{equation}
v_t = \alpha_t \epsilon - \sigma_t z_0 .
\end{equation}
Given a noisy latent $z_t \in \mathbb{R}^{T \times J \times D}$, a text embedding $c$, and a timestep embedding $t$, the denoiser predicts
\begin{equation}
\hat{v}_t = v_\theta\big(z_t, t, c\big),
\label{eq:vel_pred}
\end{equation}
where $T$, $J$, and $D$ denote the number of frames, joints, and feature channels, respectively, while $v_\theta$ denotes the denoising network.
With a noisy latent $z_t$ and its predicted velocity $\hat{v}_t$, the clean latent can be reconstructed as
\begin{equation}
\hat{z}_0 = \alpha_t z_t - \sigma_t \hat{v}_t.
\label{eq:z0_from_v}
\end{equation}
This parameterization improves training stability and provides a unified formulation encompassing both noise and data prediction.

\subsection{LoRA}
\label{sec:lora}
LoRA \citep{hu2022lora} is a parameter-efficient fine-tuning method that adapts pretrained models by injecting low-rank updates into existing weight matrices while keeping the original parameters frozen.
Given a linear projection with weight matrix $W \in \mathbb{R}^{d \times k}$, where $d$ and $k$ denote the feature dimensions of the projection, LoRA introduces a learnable low-rank update
\begin{equation}
\Delta W = \frac{\alpha}{r} B A,
\quad
A \in \mathbb{R}^{r \times k}, \;
B \in \mathbb{R}^{d \times r},
\end{equation}
where $r \ll \min(d, k)$ is the rank of the adaptation, $A$ and $B$ are learnable low-rank matrices, and $\alpha$ is a scaling factor controlling the update magnitude.
The adapted weight is given by
\begin{equation}
W' = W + \Delta W.
\end{equation}
This formulation allows efficient conditioning or adaptation of large pretrained models with a small number of additional parameters, significantly reducing training cost while not increasing the inference latency. As a result, LoRA has been successfully applied to pretrained models across various domains such as language \citep{hu2022lora}, image \citep{ryu2023low}, and motion \citep{sawdayee2025dance}.
\section{Method}
\label{sec:method}
\begin{figure*}[t]
  \centering
  \includegraphics[width=\linewidth]{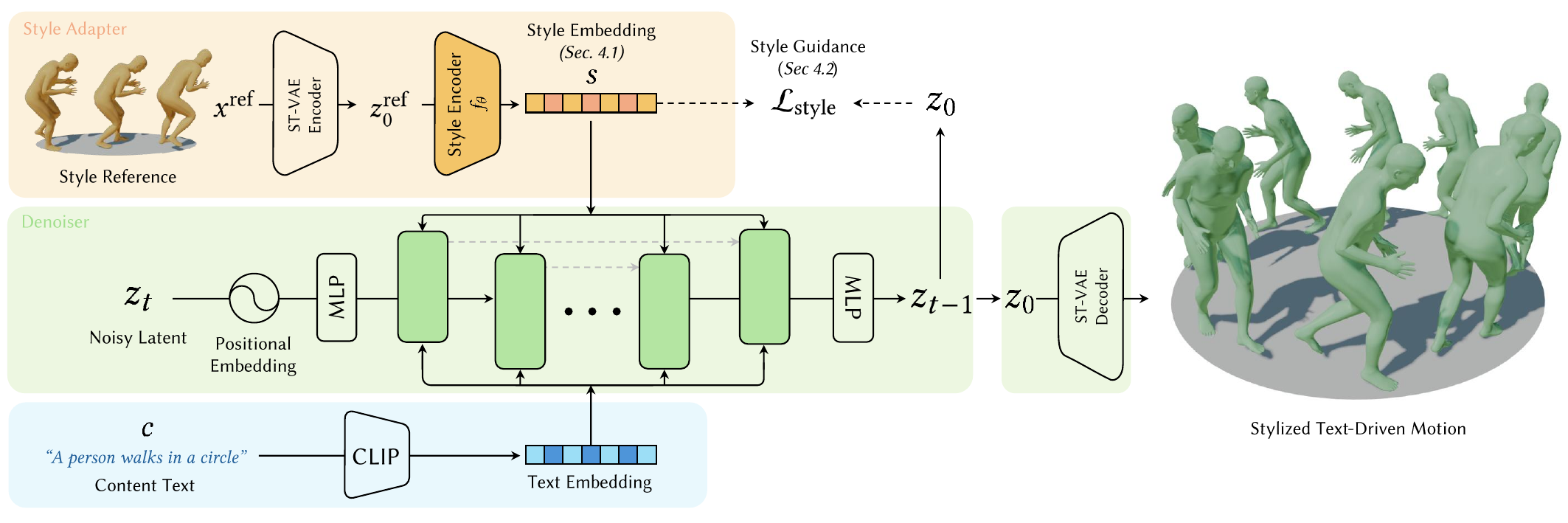}
  \caption{{Method overview}. Our method conditions motion generation on a text prompt for content and a reference motion sequence for style, producing motions that jointly reflect both content and style.}
  \label{fig:overview}
\end{figure*}

We present a style conditioning framework that leverages a pretrained text-to-motion diffusion model (SALAD~\citep{hong2025salad}) to generate stylized motion. Given a reference motion sequence, a style adapter extracts a style embedding and injects it into the denoising process using a hypernetwork-driven LoRA mechanism, enabling stylistic control while preserving the motion content specified by text (\cref{sec:style_adapter}).
Furthermore, we introduce a style-guidance mechanism at inference time that explicitly steers the denoising trajectory toward motions consistent with the reference style~(\cref{sec:style_guidance}). This guidance operates directly on the diffusion latent and complements the proposed style conditioning. The overall pipeline is shown in~\cref{fig:overview}.

\subsection{Style Adapter}
\label{sec:style_adapter}

To obtain a compact latent representation compatible with SALAD, we encode a reference motion sequence
$x^{\mathrm{ref}} \in \mathbb{R}^{T \times 263}$ into a structured motion latent representation
$z^{\mathrm{ref}}_0 \in \mathbb{R}^{T \times J \times D}$
using the ST-VAE encoder of SALAD. A style encoder $f_\theta$, implemented as a stack of MLP blocks followed by mean pooling across the joint and temporal dimensions, then extracts a global style embedding $s\in\mathbb{R}^{D}$:
\begin{equation}
s = f_\theta(z^{\mathrm{ref}}_0),
\end{equation}
which captures the spatial and temporal dynamics of motion style. Notably, we encourage style latent representations derived from different motions that share the same style to cluster in a shared region of the latent space using a supervised contrastive loss~\cite{khosla2020supervised}. This enables constructing a discriminative style latent space while remaining agnostic to the reference motion content.

\paragraph{Style-Conditioned FiLM Modulation.}
Our goal is to train a hypernetwork $f_\psi(s)$ that maps the style embedding to style-conditioned low-rank parameters that modulate intermediate features of the denoising transformer. Therefore, the network parametrization defined in~\cref{eq:vel_pred} can be reformulated as follows:
\begin{equation}
    \hat{v}_t = v_\theta\big(z_t, t, c;\, f_\psi(s)\big).
\end{equation}
The denoising transformer of SALAD incorporates feature-wise linear modulation (FiLM)~\citep{perez2018film} layers to condition intermediate features on diffusion timestep embeddings. Given a timestep embedding $e_t\in\mathbb{R}^{D}$, the pretrained FiLM generator produces
scale and shift parameters $(\gamma_0, \beta_0)$:
\begin{equation}
(\gamma_0, \beta_0) = \mathrm{split}\big(W \phi(e_t) + b\big),
\end{equation}
where $W$ and $b$ denote the weight and bias parameters of the original FiLM projection, respectively, with SiLU activation~\cite{elfwing2018silu} $\phi(\cdot)$. These parameters modulate intermediate features $H$ of the denoising transformer as follows:
\begin{equation}
H' = \gamma_0 \odot H + \beta_0,
\end{equation}
where $\odot$ denotes the element-wise multiplication.

\begin{figure}[t]
  \centering
  \includegraphics[width=1\linewidth]{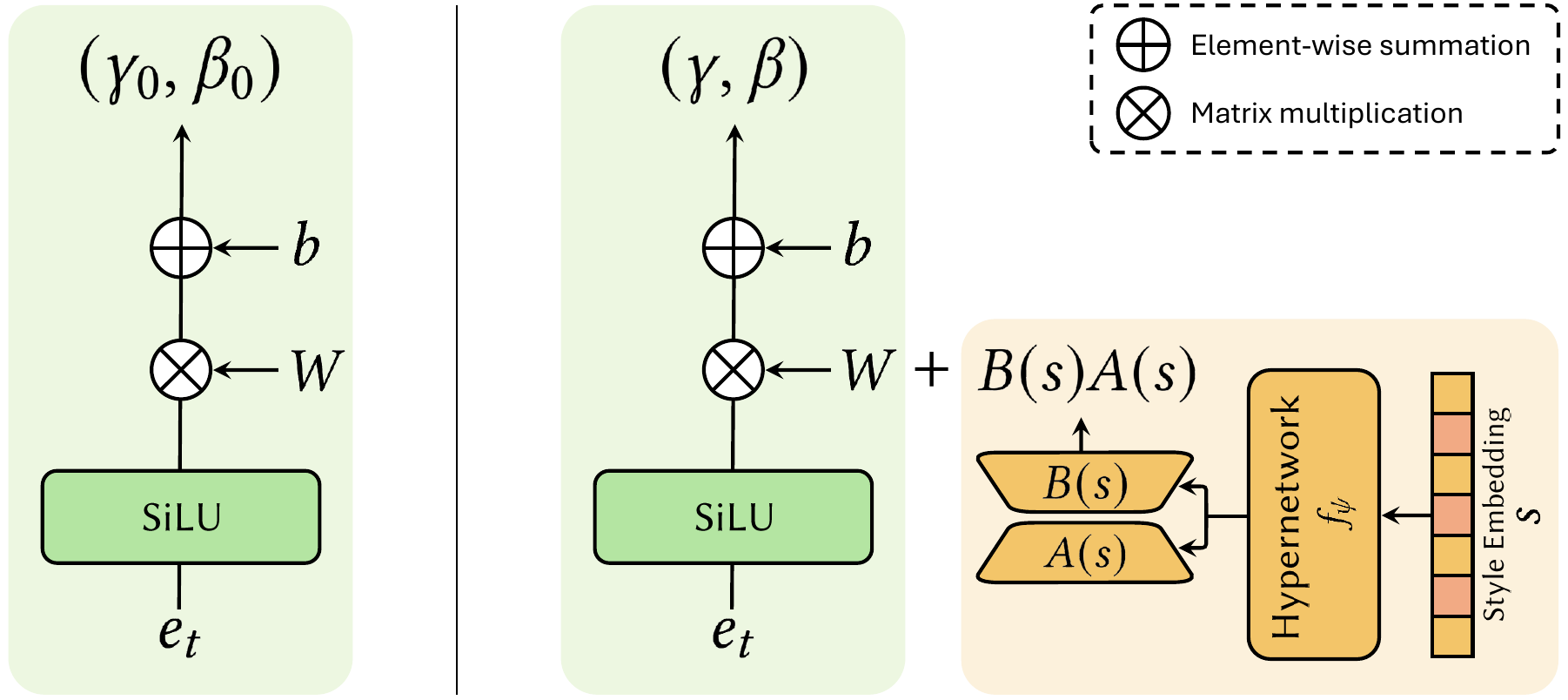}
  \caption{{Comparison between the FiLM mechanism of SALAD~\citep{hong2025salad}~(left) and our HyperLoRA~(right)}.}
  \label{fig:hyperlora}
\end{figure}

Similar to the timestep embedding that globally modulates intermediate features, motion style should be reflected throughout the entire motion sequence. Therefore, we inject motion style by applying a style-conditioned low-rank update to the pretrained FiLM generators as shown in~\cref{fig:hyperlora}. Specifically, we compute an intermediate activation $\phi(e_t)$ and augment the FiLM output with a low-rank update generated by a hypernetwork conditioned on the style embedding $s$ as follows:
\begin{equation}
(A(s), B(s)) = \mathrm{split}\big(f_\psi(s)\big).
\end{equation}
Consequently, the final FiLM parameters are obtained as
\begin{equation}
(\gamma, \beta) = \mathrm{split}\big( (W+B(s)A(s))\phi(e_t)+b \big), \\
\end{equation}
which are subsequently used to modulate intermediate features throughout the denoising process.

This design enables style-dependent modulation while preserving the original timestep conditioning and keeping the pretrained diffusion backbone frozen. By localizing adaptation to FiLM parameters, the proposed style adapter maintains the text-conditioned motion semantics learned by SALAD while providing flexible and parameter-efficient control over stylistic attributes.
Furthermore, because both $A(s)$ and $B(s)$ do not change once they are derived from a reference motion, we update the weight matrices of every FiLM layer before the denoising process so that it does not introduce latency during inference.

\subsection{Style Guidance}
\label{sec:style_guidance}

During inference, we employ two complementary guidance mechanisms to control both semantic correctness and stylistic fidelity of the generated motion. Classifier-free guidance \citep{ho2022classifier} is used to balance unconditional, text-conditioned, and style-conditioned predictions at the model output level, while an additional style encoder guidance mechanism directly steers the denoising trajectory using gradients derived from a learned style representation. Together, these mechanisms provide fine-grained control over motion generation without retraining the diffusion model.

\paragraph{Classifier-Free Guidance}
Following SMooDi~\cite{zhong2024smoodi}, we adopt classifier-free guidance to balance unconditional, text-conditioned, and style-conditioned predictions. Specifically, we compute three velocity predictions corresponding to unconditional, text-conditioned, and style-conditioned inputs, and combine them as
\begin{equation}
\hat{v}_t =
\hat{v}_t^{\mathrm{uncond}}
+ w_{\mathrm{text}}\big(\hat{v}_t^{\mathrm{text}} - \hat{v}_t^{\mathrm{uncond}}\big)
+ w_{\mathrm{style}}\big(\hat{v}_t^{\mathrm{style}} - \hat{v}_t^{\mathrm{text}}\big),
\end{equation}
where $w_{\mathrm{text}}$ and $w_{\mathrm{style}}$ control the strength of text and style guidance, respectively. This formulation extends standard classifier-free guidance by explicitly incorporating style-conditioned predictions, enabling simultaneous control over motion semantics and stylistic attributes.

\paragraph{Style Encoder Guidance}
While classifier-free guidance modulates the denoiser output, we further enhance stylistic fidelity by introducing a gradient-based guidance mechanism driven by a learned style encoder. Specifically, we leverage the trained style encoder to formulate a style guidance objective term, as its discriminative training enables continuous and robust style supervision even for unseen styles. This approach is inspired by classifier guidance \citep{dhariwal2021diffusion}, but differs in that guidance signals are derived from continuous style embeddings extracted from reference motion rather than discrete style classifiers.

Given the predicted clean latent $\hat{z}_0$ at timestep $t$ using~\cref{eq:z0_from_v}, we derive the predicted style embedding $\hat{s}_t$ using the style encoder $f_\theta$:
\begin{equation}
\hat{s}_t = f_\theta(\hat{z}_0).
\end{equation}
We then define a style consistency objective that encourages the generated motion to match the reference style embedding $s$:
\begin{equation}
\mathcal{L}_{\mathrm{style}}
= \left\lVert \hat{s}_t - s \right\rVert_2^2 .
\label{eq:style_loss}
\end{equation}
This objective aims to incorporate the learned representation of the style encoder, which captures style features in a content-agnostic manner and maps motions sharing the same style to a shared representation.

The gradient of this objective is backpropagated to the noisy latent $z_t$, yielding a style guidance signal,
\begin{equation}
    g_t = \nabla_{z_t} \mathcal{L}_{\mathrm{style}},
\end{equation}
which is used to steer the denoising trajectory toward motions that better preserve stylistic attributes of the reference motion. Taking inspiration from gradient-based guided diffusion~\citep{dhariwal2021diffusion}, we apply a normalized gradient step on the noisy latent $z_t$ prior to the
reverse diffusion update:
\begin{equation}
z_t^{\mathrm{guided}} =
z_t - \lambda_{\mathrm{style}}
\frac{g_t}{\lVert g_t \rVert_2 + \epsilon},
\end{equation}
where $\lambda_{\mathrm{style}}$ controls the strength of the style guidance and $\epsilon$ is a small constant for numerical stability. The reverse diffusion step is then performed using $(z_t^{\mathrm{guided}}, \hat{v}_t)$.

This formulation enables continuous control over the influence of style during generation, allowing a smooth trade-off between text-conditioned motion content and reference-driven stylistic expression. Notably, style encoder guidance does not rely on discrete style classifiers or predefined style categories, making it applicable to unseen styles. In practice, this guidance mechanism complements classifier-free guidance by reinforcing stylistic consistency at inference time while preserving semantic alignment imposed by text conditioning.

\subsection{Training}
\label{sec:training}

During training, we freeze the pretrained denoiser of SALAD and jointly train only the style encoder $f_\theta$ and the hypernetwork $f_\psi$ using the 100STYLE dataset~\citep{mason2022real} retargeted to the SMPL skeleton by \citeN{zhong2024smoodi}. While SMooDi~\citep{zhong2024smoodi} utilizes MotionGPT-generated~\citep{jiang2023motiongpt} text descriptions for training, we observed that such descriptions often exhibit low semantic alignment with the motion sequences, hindering the training process. Instead, we pair each motion sequence with simple, manually defined text descriptions based on locomotion categories (e.g., "\textit{A person walks forward.}") for semantically consistent conditioning signals. 
Furthermore, to encourage content-invariant style representations, we intentionally misalign motion content by randomly pairing text prompts with style reference motions of the same style but different content during training.

The model is optimized under the velocity prediction objective
\begin{equation}
\mathcal{L}_{\mathrm{vel}} 
= \mathbb{E}_{z_0, t, \epsilon}\!\left[
\lVert \hat{v}_t - v_t \rVert_2^2
\right],
\label{eq:vel_loss}
\end{equation}
which measures the discrepancy between predicted and true velocities across timesteps. Additionally, to introduce semantic structure to the style embedding space and improve generalization to unseen styles, we employ a supervised contrastive loss~\citep{khosla2020supervised} on the extracted style embeddings. Given style embeddings $\{s_i\}$ and style labels $\{y_i\}$, we define for each anchor $i$ the positive and negative sets as $\mathcal{P}(i)=\{p \neq i: y_p = y_i\}$ and $\mathcal{A}(i)=\{a \neq i\}$, respectively. The supervised contrastive loss is defined as:
\begin{equation}
\begin{aligned}
\mathcal{L}_{\mathrm{supcon}}
= \sum_i \frac{-1}{|\mathcal{P}(i)|}
\sum_{p\in\mathcal{P}(i)}
\log
\frac{\exp\!\big(\cos(s_i,s_p)/\tau\big)}
{\sum_{a\in\mathcal{A}(i)} \exp\!\big(\cos(s_i,s_a)/\tau\big)},
\end{aligned}
\label{eq:supcon}
\end{equation}
where $\tau$ denotes the temperature parameter. This objective pulls together embeddings of the same style while pushing apart those of different styles regardless of the motion content, forming a semantically structured style manifold.

The overall training objective is expressed as
\begin{equation}
\mathcal{L}_{\mathrm{total}}
= 
\mathcal{L}_{\mathrm{vel}}
+ \lambda_{\mathrm{supcon}}\,\mathcal{L}_{\mathrm{supcon}},
\label{eq:total_loss}
\end{equation}
where $\lambda_{\mathrm{supcon}}$ balances the contribution of the supervised contrastive loss. This objective encourages content-preserving style modulation while promoting generalization to unseen styles.

\paragraph{Implementation Details}
Our model was trained for 100 epochs on a single NVIDIA RTX A5000 GPU with 24GB VRAM, which required 2.47 hours.
We used the Adam optimizer~\protect\citep{kingma2017adammethodstochasticoptimization} with a fixed learning rate of $1\mathrm{e}{-4}$ and a batch size of 64.
We set $w_\mathrm{text}$, $w_\mathrm{style}$, $\lambda_\mathrm{style}$, $\tau$, and $\lambda_\mathrm{supcon}$ as 7.5, 1.5, 0.75, 0.07, and 1, respectively.
\begin{table*}[t]
  \caption{{Quantitative comparison of baselines.}
  $\uparrow$ and $\downarrow$ indicate that higher and lower values are better, respectively. \textbf{Bold}, \underline{underline}, and \textit{italic} represent the first-, second-, and third-best results, respectively.}
  \label{tab:comparison_quan}
  \centering
  \small
  \setlength{\tabcolsep}{6pt}
  \begin{tabular*}{\textwidth}{@{\extracolsep{\fill}} l c c c c c}
    \toprule
    & \multicolumn{1}{c}{\textbf{Style Expression}}
    & \multicolumn{2}{c}{\textbf{Content Preservation}}
    & \multicolumn{2}{c}{\textbf{Motion Quality}} \\
    \cmidrule(lr){2-2} \cmidrule(lr){3-4} \cmidrule(lr){5-6}
    \textbf{Method} 
      & \textbf{SRA (\%)} $\uparrow$
      & \textbf{R-Precision} $\uparrow$
      & \textbf{MM Dist} $\downarrow$
      & \textbf{FID} $\downarrow$
      & \textbf{FSR} $\downarrow$ \\
    \midrule
    SMooDi~\citep{zhong2024smoodi} 
      & {\underline{75.818}} & 0.571 & 4.477 & 1.609 & 0.124 \\
    LoRA-MDM~\citep{sawdayee2025dance} 
      & 42.047 & \textbf{0.772} & \textbf{3.234} & \textbf{0.607} & \textbf{0.049} \\
    \citeN{wu2025semantically}
      & {\textit{56.509}} & {0.623} & {4.222} & {2.987} & {0.134} \\
    Ours
      & \textbf{76.034} & \textit{0.721} & \textit{3.519} & {1.138} & \textit{0.086} \\
    \midrule
    SMooDi \textit{w/o} classifier guidance
      & {43.417} & 0.630 & 4.085 & \textit{1.050} & 0.111 \\
    Ours \textit{w/o} style encoder guidance
      & {56.336} & \underline{0.738} & \underline{3.379} & \underline{0.617} & \underline{0.071} \\
    \bottomrule
  \end{tabular*}
\end{table*}

\begin{figure*}[t]
  \centering
  \includegraphics[width=\linewidth]{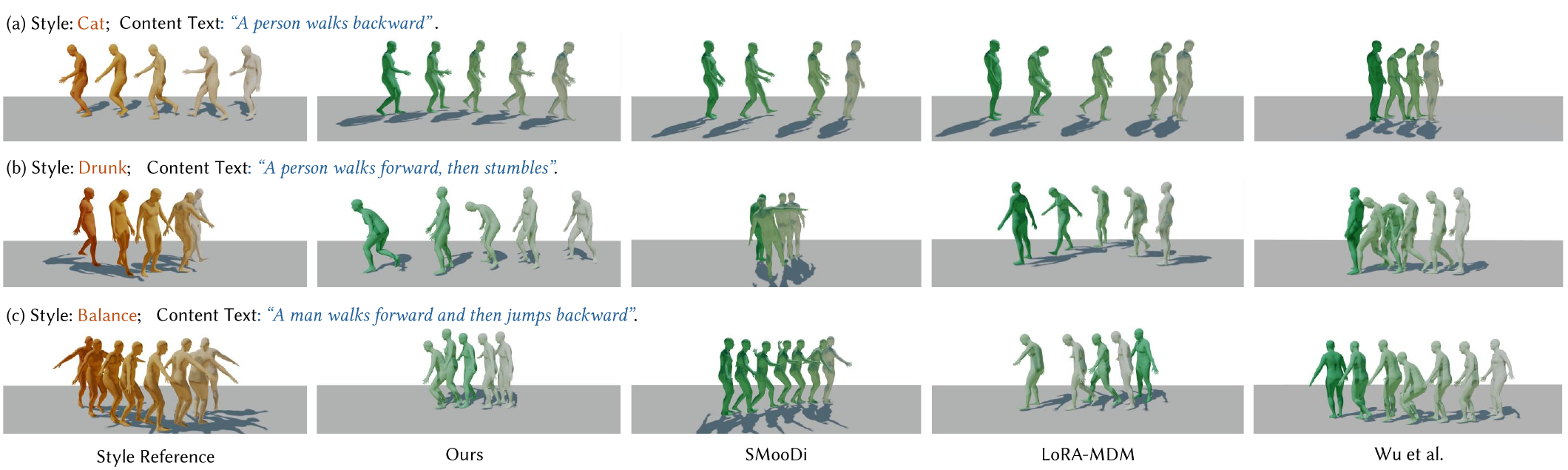}
  \caption{{Qualitative evaluation}. We present stylized motion sequences generated with three different style reference and context description pairs. Brighter colors indicate earlier frames.}
  \label{fig:comparison}
\end{figure*}

\section{Experiments}
\label{sec:experiments}

\subsection{Settings}
\label{sec:Settings}

\paragraph{Dataset}
We conducted experiments using the HumanML3D dataset \citep{guo2022generating} and the 100STYLE dataset~\citep{mason2022real}. We used the 100STYLE dataset retargeted to the SMPL skeleton by \citeN{zhong2024smoodi}. During generation, we used textual descriptions from HumanML3D as content prompts and stylized motion sequences from 100STYLE as style references.

\paragraph{Evaluation Metrics}
Following~\citeN{zhong2024smoodi}, we evaluated generated motions across three categories: style expression, content preservation, and motion quality. Style expression is quantified using Style Recognition Accuracy (SRA), computed with a pretrained style classifier. Content preservation is evaluated using motion–text retrieval metrics, R-Precision and Multimodal Distance (MM Dist). We report SRA as Top-5 accuracy and R-Precision as Top-3 accuracy. Motion quality is assessed using Fréchet Inception Distance (FID) and Foot Skating Ratio (FSR). In addition, we conducted a user study to evaluate the perceptual naturalness of the generated results.

\paragraph{Baselines}
We compared our approach against three representative methods that condition motion generation with style: SMooDi \citep{zhong2024smoodi}, LoRA-MDM~\citep{sawdayee2025dance}, and~\citeN{wu2025semantically}. SMooDi employs a ControlNet-based~\citep{zhang2023adding} style adapter combined with classifier-based and classifier-free guidance. In contrast, LoRA-MDM trains an independent LoRA adapter~\cite{hu2022lora} for each style. \citeN{wu2025semantically} explores unsupervised style modeling within a diffusion-based framework.

\subsection{Quantitative Comparison}
A quantitative comparison is presented in \cref{tab:comparison_quan}. Our method achieved the highest SRA, indicating superior style expression compared to the baselines. Among the baseline methods, SMooDi demonstrated competitive style expression but underperformed in content preservation and motion quality. In contrast, LoRA-MDM maintained strong content preservation and motion quality at the cost of substantially reduced style expression. \citeN{wu2025semantically} demonstrated strong style expression without relying on explicit style guidance, but showed comparatively weaker performance in content preservation and motion quality. While the baseline results indicate a clear trade-off between style expression and motion fidelity, our method demonstrates robust style expression with only a minimal compromise, maintaining competitive performance in content preservation and motion quality. Notably, our method outperformed SMooDi in SRA although SMooDi employs the same pretrained style classifier used for evaluation as a guidance signal during inference, highlighting the effectiveness of our learned style representation.

We additionally report results without style guidance for both methods to evaluate their intrinsic performance. Even without guidance, our model demonstrated competitive style expression compared to other unguided methods while not sacrificing other metric values. These results indicate that our HyperLoRA-based style adapter captures style features effectively while not compromising the text fidelity and motion quality of the pretrained backbone.

\begin{table}[t]
  \caption{{User Study.}
  Subjective comparison of our method with baselines across style expression, content preservation, and motion quality (mean $\pm$ CI$_{95}$ margin).}
  \label{tab:user_study}
  \centering
  \scriptsize
  \setlength{\tabcolsep}{2pt}
  \renewcommand{\arraystretch}{1.05}
  \begin{tabular*}{\columnwidth}{@{\extracolsep{\fill}} l c c c}
    \toprule
    \textbf{Method}
      & \textbf{Style Expression}
      & \textbf{Content Preservation}
      & \textbf{Motion Quality} \\
    \midrule
    SMooDi
    & {3.195 $\pm$ 0.243}
    & {3.085 $\pm$ 0.173}
    & {2.705 $\pm$ 0.171} \\
    LoRA-MDM
    & {3.035 $\pm$ 0.272}
    & {3.765 $\pm$ 0.234}
    & {3.445 $\pm$ 0.269} \\
    \citeN{wu2025semantically}
    & {3.090 $\pm$ 0.288}
    & {3.180 $\pm$ 0.230}
    & {2.765 $\pm$ 0.240} \\
    Ours
    & {\textbf{4.215} $\pm$ 0.161}
    & {\textbf{4.520} $\pm$ 0.172}
    & {\textbf{4.050} $\pm$ 0.194} \\
    \bottomrule
  \end{tabular*}
\end{table}
\begin{table*}[!t]
  \caption{{Ablation studies on supervised contrastive loss and style encoder guidance across four variants of our model with different training data distributions. The first column indicates the number of styles used for training.}}
  \label{tab:ablation}
  \centering
  \small
  \setlength{\tabcolsep}{6pt}

  \begin{tabular*}{\textwidth}{@{\extracolsep{\fill}} c c c c c c c c c}
    \toprule
    & & & & \multicolumn{1}{c}{\textbf{Style Expression}}
      & \multicolumn{2}{c}{\textbf{Content Preservation}}
      & \multicolumn{2}{c}{\textbf{Motion Quality}} \\
    \cmidrule(lr){5-5} \cmidrule(lr){6-7} \cmidrule(lr){8-9}
    \textbf{Styles}
      & \textbf{Setting}
      & $\mathcal{L}_\mathrm{supcon}$
      & Style Encoder Guidance
      & \textbf{SRA (\%)} $\uparrow$
      & \textbf{R-Prec} $\uparrow$
      & \textbf{MM Dist} $\downarrow$
      & \textbf{FID} $\downarrow$
      & \textbf{FSR} $\downarrow$ \\
    \midrule

    \textbf{100}
      & (a) & $\bigcirc$ & $\bigcirc$
      & \underline{76.034} & 0.721 & \textit{3.519} & \textit{1.138} & \textit{0.086} \\
      & (b) & $\times$ & $\bigcirc$ 
      & \textbf{77.888} & \textit{0.722} & 3.589 & 1.454 & 0.093 \\
      & (c) & $\bigcirc$ & $\times$
      & \textit{56.336} & \underline{0.738} & \underline{3.379} & \underline{0.617} & \underline{0.071} \\
      & (d) & $\times$ & $\times$
      & 53.211 & \textbf{0.745} & \textbf{3.361} & \textbf{0.570} & \textbf{0.069} \\
    \midrule

    \textbf{75}
      & (a) & $\bigcirc$ & $\bigcirc$
      & \underline{72.004} & \underline{0.726} & \textit{3.523} & \textit{1.090} & \textit{0.094} \\
      & (b) & $\times$ & $\bigcirc$ 
      & \textbf{74.009} & 0.712 & 3.630 & 1.426 & 0.101 \\
      & (c) & $\bigcirc$ & $\times$
      & \textit{50.539} & \textbf{0.746} & \textbf{3.375} & \textbf{0.596} & \textbf{0.076} \\
      & (d) & $\times$ & $\times$
      & 48.125 & \textbf{0.746} & \underline{3.382} & \underline{0.631} & \underline{0.079} \\
    \midrule

    \textbf{50}
      & (a) & $\bigcirc$ & $\bigcirc$
      & \textbf{63.082} & \textit{0.745} & \textit{3.399} & \textit{0.887} & \textit{0.087} \\
      & (b) & $\times$ & $\bigcirc$ 
      & \underline{62.996} & 0.739 & 3.453 & 1.101 & 0.095 \\
      & (c) & $\bigcirc$ & $\times$
      & \textit{44.741} & \underline{0.747} & \underline{3.335} & \underline{0.588} & \textbf{0.075} \\
      & (d) & $\times$ & $\times$
      & 41.401 & \textbf{0.753} & \textbf{3.325} & \textbf{0.585} & \underline{0.077} \\
    \midrule

    \textbf{25}
      & (a) & $\bigcirc$ & $\bigcirc$
      & \textbf{52.716} & \underline{0.757} & \underline{3.321} & \textit{1.038} & \textit{0.087} \\
      & (b) & $\times$ & $\bigcirc$ 
      & \underline{48.254} & \textbf{0.765} & 3.348 & 1.171 & 0.090 \\
      & (c) & $\bigcirc$ & $\times$
      & \textit{36.034} & 0.751 & \textit{3.337} & \underline{0.765} & \underline{0.082} \\
      & (d) & $\times$ & $\times$
      & 32.198 & \textit{0.755} & \textbf{3.316} & \textbf{0.700} & \textbf{0.080} \\
    \bottomrule
  \end{tabular*}
  
\end{table*}

\paragraph{User Study}
To account for the subjective nature of style, we conducted a user study with human evaluators. Specifically, we asked participants to rate a series of generated motions on a five-point Likert scale for style expression, content preservation, and motion quality, with five being the best. We recruited 20 participants (10 males and 10 females; aged 23--35), who were presented with a total of 40 videos (10 results $\times$ 4 methods) in a randomized order. The length of each video was shorter than 10 seconds.
As shown in~\cref{tab:user_study}, our method significantly outperformed all baselines by a clear margin across all evaluation criteria. Especially, participants rated our method substantially higher in style expression, indicating more expressive and recognizable stylistic cues. While SMooDi and \citeN{wu2025semantically} showed comparable style expression to LoRA-MDM, they fell behind in content preservation and motion quality, consistent with their inferior quantitative performance reported in~\cref{tab:comparison_quan}. Overall, these results demonstrate that the proposed method effectively balances expressive stylization with motion fidelity in human perception.

\subsection{Qualitative Comparison}
A qualitative comparison of the four methods is presented in \cref{fig:comparison}.
Overall, our method produced more expressive and semantically consistent motions than the baselines, effectively preserving both stylistic cues and content specified by text. In case (a), ours preserved swinging hand movements during walking as observed in the style reference, whereas SMooDi, LoRA-MDM, and \citeN{wu2025semantically} failed to retain these stylistic details despite maintaining text faithfulness. In case (b), SMooDi and LoRA-MDM exhibited style expressions but failed to accurately reflect the content text (i.e., \emph{stumbles}), while \citeN{wu2025semantically} captured the content text but failed to preserve the style. In contrast to the baselines, ours successfully maintained both style and content fidelity. In case (c), the baseline methods failed to reflect the intended content despite favorable style expression. These qualitative results demonstrate that our method better balances expressive stylization and content preservation despite not relying on a pretrained style classifier or style-specific fine-tuning. Animated results are presented in the supplementary video.

\begin{figure*}[t]
  \centering
  \includegraphics[width=\linewidth]{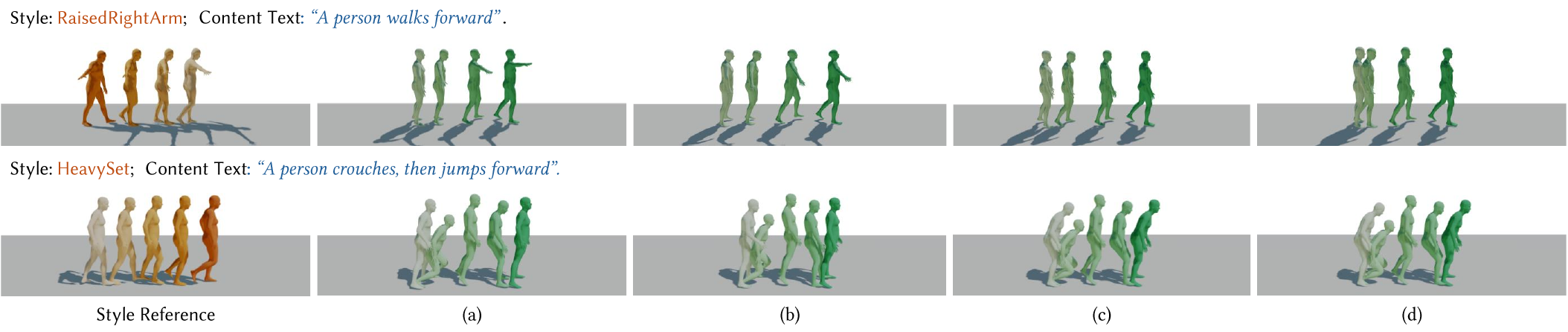}
  \caption{{Ablation study results on the supervised contrastive loss and the style encoder guidance when trained on 25 styles.} (a)--(d) annotations follow those in~\cref{tab:ablation}: (a) full, (b) without $\mathcal{L}_\mathrm{supcon}$, (c) without style encoder guidance, and (d) without both. Brighter colors indicate earlier frames.}
  \label{fig:ablation}
\end{figure*}

\subsection{Ablation Study}
\label{sec:ablation}
We conducted ablation studies on three key components of our method: the supervised contrastive loss, style encoder guidance, and the style-conditioning architecture. To evaluate the generalization of our method to unseen styles, we additionally report results on model variants trained with subsets of the style dataset, containing only 75, 50, and 25 styles during training.

\paragraph{Supervised Contrastive Loss}
We compared our models to variants trained without the supervised contrastive loss, while still using the style encoder guidance although it does not reliably disentangle stylistic attributes from the motion content. As shown in (a) and (b) of~\cref{tab:ablation}, when trained on all 100 styles, removing $\mathcal{L}_\mathrm{supcon}$ resulted in a slight improvement in SRA albeit with a significant degradation in FID. However, as the number of available training styles decreases, the benefit of the supervised contrastive loss becomes more pronounced, as evidenced by our model's highest SRA for 50- and 25- style settings. In addition, our model consistently maintained better motion quality than the variant without $\mathcal{L}_{\text{supcon}}$, as evidenced by the lower FID and FSR. These results indicate that the supervised contrastive loss encourages a structured and discriminative style representation that is independent from the motion content, which contributes to improving robustness and generalization to unseen styles, particularly when the training data is limited.
These observations align with the qualitative results shown in the top row of (a) and (b) of~\cref{fig:ablation}, where the absence of $\mathcal{L}_\mathrm{supcon}$ led to less expressive stylization with arms lower than those in the style reference.
We also compared the effectiveness of $\mathcal{L}_\mathrm{supcon}$ when the style encoder guidance was not employed. As shown in (c) and (d) of~\cref{tab:ablation}, using $\mathcal{L}_\mathrm{supcon}$ consistently achieved better SRA for all style settings while not compromising content preservation and motion quality. These results demonstrate that the supervised contrastive loss is crucial to improving the intrinsic model performance.

\begin{figure}[t]
  \centering
  \includegraphics[width=\linewidth]{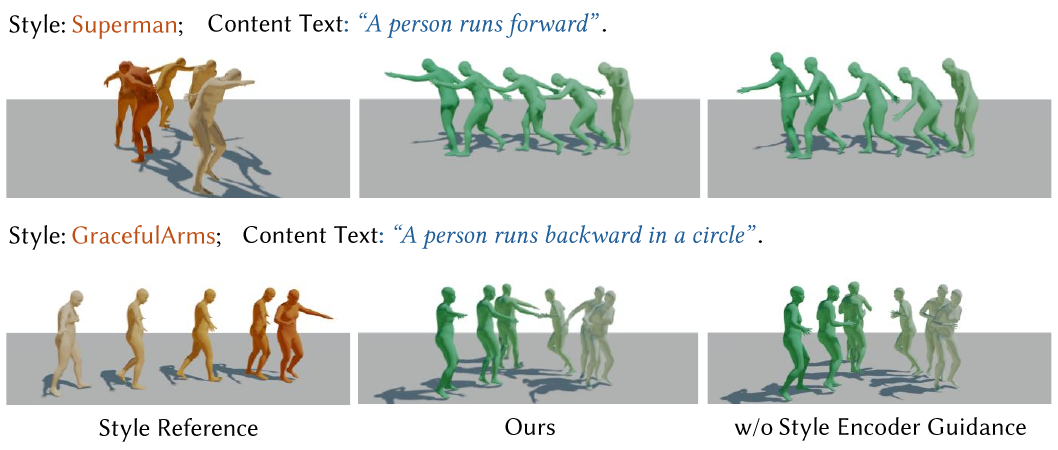}
  \caption{{Ablation study results on the style encoder guidance.} Brighter colors indicate earlier frames.}
  \label{fig:ablation_guidance}
\end{figure}

\paragraph{Style Encoder Guidance}
To assess the impact of the style encoder guidance, we evaluated variants of our method without this component. As shown in (a) and (c) of \cref{tab:ablation}, removing the style encoder guidance led to a substantial drop in SRA across all settings, confirming that inference-time style guidance plays a critical role in achieving strong style expression. This effect is further supported by qualitative results: \cref{fig:ablation_guidance} shows the benefit of style encoder guidance for a model trained on all 100 styles, while case (a) and (c) of \cref{fig:ablation} show that the absence of style encoder guidance significantly compromised style expression for the model trained with only 25 styles. Together, these results demonstrate that style encoder guidance consistently enhances style expression regardless of the number of styles available during training. At the same time, we observed that removing style encoder guidance consistently improved content preservation and motion quality metric scores except R-Precision and MM Dist on 25-style samples. This highlights a trade-off between expressive stylization and motion fidelity, which is consistent with prior observations in classifier-guided methods such as SMooDi where the guidance improves style expression at the expense of motion realism.

\paragraph{Architecture}
To evaluate the effectiveness of our HyperLoRA-based style adapter against a widely adopted stylization architecture, we compared our method with a ControlNet-based variant integrated into the same diffusion backbone. As shown in \cref{tab:ablation_arch}, the ControlNet-based variant struggled to adapt style into the generation process, resulting in low SRA despite strong performance in R-Precision and FID, indicating its limited capacity to inject stylistic attributes to the motion diffusion model. In contrast, our method achieved significantly higher SRA and lower inference time while maintaining competitive R-Precision and FID, demonstrating superior overall stylization performance with favorable content preservation and motion quality.

\begin{table}[t]
  \caption{{Comparison of architectural performance.}
  We compared our HyperLoRA-based style adapter with a ControlNet-style adapter integrated into the same backbone.}
  \label{tab:ablation_arch}
  \centering
  \footnotesize
  \setlength{\tabcolsep}{4pt}
  \renewcommand{\arraystretch}{1.1}
  \begin{tabular*}{\columnwidth}{@{\extracolsep{\fill}} l c c c c}
    \toprule
    \textbf{Method}
      & \textbf{SRA (\%)} $\uparrow$
      & \textbf{R-Prec} $\uparrow$
      & \textbf{FID} $\downarrow$
      & \textbf{Inference Time (ms)} $\downarrow$ \\
    \midrule
    Ours
      & \textbf{76.034}
      & 0.721
      & 1.138
      & {6156.119} \\
    \quad {\textit{w/o} guidance}
      & {56.336}
      & {0.738}
      & {0.617}
      & {\textbf{5999.534}} \\
    \midrule
    ControlNet-based
      & 32.069
      & {0.802}
      & {0.405}
      & {6918.383} \\
    \quad {\textit{w/o} guidance}
      & {12.112}
      & {\textbf{0.860}}
      & {\textbf{0.059}}
      & {6846.849} \\
    \bottomrule
  \end{tabular*}
\end{table}

\subsection{Generalization to Unseen Styles}
\begin{figure}[t]
  \centering
  \includegraphics[width=\linewidth]{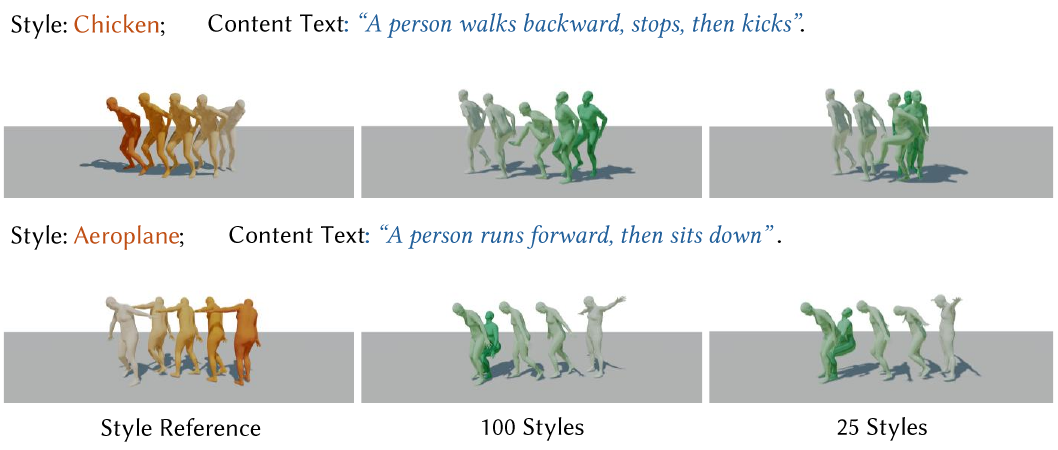}
  \caption{{Comparison of motions generated using models trained on 100 styles and 25 styles. Reference styles are unseen in the 25-style setting. Brighter colors indicate earlier frames.}}
  \label{fig:ablation_styles}
\end{figure}

Our method is able to reflect motion styles that are not observed during training, enabled by the combination of supervised contrastive loss and explicit guidance from the style encoder.
To demonstrate the generalization capability to unseen styles, we compare the model variants: one trained on the full set of 100 styles, and another trained on a subset of 25 styles.
As shown in~\cref{fig:ablation_styles}, the model trained with only 25 styles was still able to generate plausible and coherent stylized motions, achieving performance comparable to the model trained on the full style set.
These results indicate that our proposed approach exhibits strong robustness and generalization to unseen motion styles.

\section{Applications}
\label{sec:Applications}
Beyond stylized text-to-motion generation, our framework supports a range of downstream applications, including constraint-guided generation during sampling and the editing of existing motion sequences.

\subsection{Generation with Constraints}
\begin{figure}[t]
  \centering
  \includegraphics[width=\linewidth]{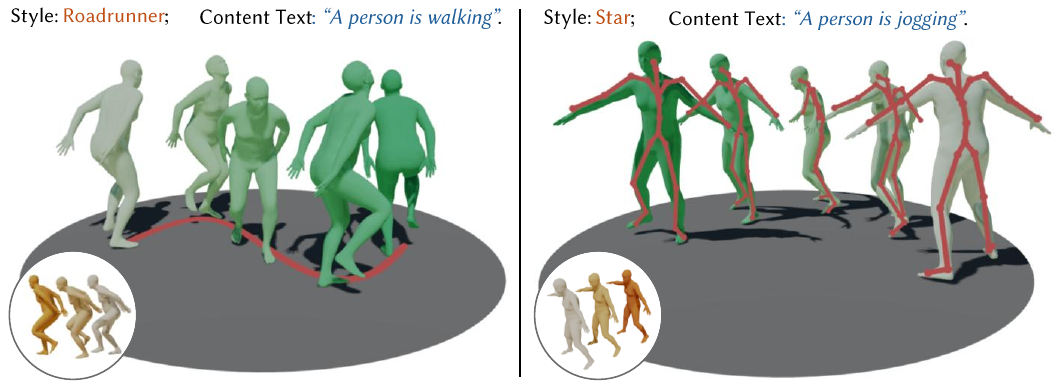}
  \caption{{Controlled stylized motion generation using trajectory (left) and keyframe (right) constraints. Brighter colors indicate earlier frames.}}
  \label{fig:motion_control}
\end{figure}

Our framework can be combined with external motion constraints such as trajectory and keyframes (\cref{fig:motion_control}), similar to prior work on controllable motion diffusion and inference-time guidance~\citep{karunratanakul2023guided, xie2023omnicontrol, karunratanakul2024optimizing}. At each denoising step, we recover the predicted clean sample from the current noisy latent, decode it into motion features, and compute constraint losses on the decoded motion. The resulting gradients are used to update the latent during sampling, so that the generated motion progressively conforms to the control signal. Specifically, trajectory guidance minimizes the discrepancy between the generated and target root trajectory, while keyframe guidance minimizes the distance between generated and target joint configurations at sparse control poses. Because these control objectives are applied together with style guidance throughout sampling, the generated motion can follow the desired trajectory or keyframes while still retaining the target style.

\subsection{Motion Style Transfer}
\begin{figure}[t]
  \centering
  \includegraphics[width=\linewidth]{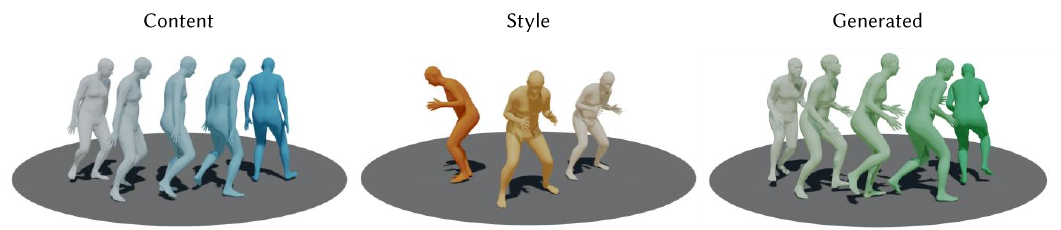}
  \caption{{Motion style transfer. A content motion input (left) is combined with a style reference (middle) to generate a stylized motion sequence (right). Brighter colors indicate earlier frames.}}
  \label{fig:motion_transfer}
\end{figure}

Beyond generation from noise, our method can edit existing motions through DDIM inversion~\citep{song2020denoising} (\cref{fig:motion_transfer}), 
in line with recent diffusion-based motion editing and transfer approaches~\citep{zhong2024smoodi, raab2024monkey}. We first encode a source motion into the motion latent space and invert it to a noisy latent using the DDIM forward process. Starting from this inverted latent, we then apply our standard reverse denoising procedure under a target text condition and a target style embedding, producing a motion sequence that follows the content and style while remaining anchored to the original motion dynamics. In this way, the inverted latent serves as a source-derived structural prior, enabling motion style transfer without retraining or additional control modules.

\section{Limitations, Future Work, and Conclusion}
\label{sec:conclusion}

While our framework demonstrates expressive stylization and strong generalization, a number of limitations suggest directions for future work. Because our method is trained on the locomotion-centric 100STYLE dataset, the learned style representations are biased toward locomotion, which limits generalization to broader action categories. Moreover, although 100STYLE provides discrete style labels, style is inherently abstract and entangled with content, raising the question of whether a style representation can be meaningfully defined in isolation without a content reference. One promising direction is to define style relative to an anchor motion or content class, enabling relational representations that better disentangle stylistic attributes from motion semantics. In addition, while we impose structure on the latent style space using supervised contrastive learning to mitigate label discreteness, richer and more generalizable style representations may be learned by leveraging large unlabeled motion datasets, as explored in recent work~\citep{wu2025semantically}.

In conclusion, we introduce a hypernetwork-driven style conditioning framework for stylized text-to-motion diffusion that combines the efficiency of LoRA with the flexibility of a unified, generalizable style adapter. By dynamically predicting LoRA parameters from a reference motion and injecting them into a pretrained diffusion backbone, our approach enables expressive and scalable motion stylization without the computational overhead of ControlNet-style conditioning or the scalability limitations of per-style fine-tuning. Central to our framework is a structured latent style representation learned with supervised contrastive learning, which supports robust style extraction from arbitrary reference motions, including unseen styles, while preserving content fidelity and motion quality. Extensive experiments on HumanML3D and 100STYLE demonstrate that our method achieves state-of-the-art stylized motion generation, highlighting our method as a promising paradigm for expressive and generalizable motion style modulation.

\bibliographystyle{ACM-Reference-Format}
\bibliography{bib/main}

@String{Computer = "{IEEE} Computer" }

@String{Springer = "Springer-Verlag" }

@inproceedings{zhong2024smoodi,
  title={Smoodi: Stylized motion diffusion model},
  author={Zhong, Lei and Xie, Yiming and Jampani, Varun and Sun, Deqing and Jiang, Huaizu},
  booktitle={European Conference on Computer Vision},
  pages={405--421},
  year={2024},
  organization={Springer}
}

@inproceedings{sawdayee2025dance,
  title={Dance Like a Chicken: Low-Rank Stylization for Human Motion Diffusion},
  author={Sawdayee, Haim and Guo, Chuan and Tevet, Guy and Zhou, Bing and Wang, Jian and Bermano, Amit H},
  booktitle={Computer Graphics Forum},
  pages={e70365},
  year={2025},
  organization={Wiley Online Library}
}

@inproceedings{wu2025semantically,
  title={Semantically Consistent Text-to-Motion with Unsupervised Styles},
  author={Wu, Linjun and Tang, Xiangjun and Cong, Jingyuan and Wang, He and Hu, Bo and Gong, Xu and Li, Songnan and Liao, Yuchen and Wu, Yiqian and Liu, Chen and others},
  booktitle={Proceedings of the Special Interest Group on Computer Graphics and Interactive Techniques Conference Conference Papers},
  pages={1--10},
  year={2025}
}

@inproceedings{zhang2023adding,
  title={Adding conditional control to text-to-image diffusion models},
  author={Zhang, Lvmin and Rao, Anyi and Agrawala, Maneesh},
  booktitle={Proceedings of the IEEE/CVF international conference on computer vision},
  pages={3836--3847},
  year={2023}
}

@article{hu2022lora,
  title={Lora: Low-rank adaptation of large language models.},
  author={Hu, Edward J and Shen, Yelong and Wallis, Phillip and Allen-Zhu, Zeyuan and Li, Yuanzhi and Wang, Shean and Wang, Lu and Chen, Weizhu and others},
  journal={ICLR},
  volume={1},
  number={2},
  pages={3},
  year={2022}
}

@article{tevet2022human,
  title={Human motion diffusion model},
  author={Tevet, Guy and Raab, Sigal and Gordon, Brian and Shafir, Yonatan and Cohen-Or, Daniel and Bermano, Amit H},
  journal={arXiv preprint arXiv:2209.14916},
  year={2022}
}

@inproceedings{hong2025salad,
  title={SALAD: Skeleton-aware Latent Diffusion for Text-driven Motion Generation and Editing},
  author={Hong, Seokhyeon and Kim, Chaelin and Yoon, Serin and Nam, Junghyun and Cha, Sihun and Noh, Junyong},
  booktitle={Proceedings of the Computer Vision and Pattern Recognition Conference},
  pages={7158--7168},
  year={2025}
}

@article{zhang2024motiondiffuse,
  title={Motiondiffuse: Text-driven human motion generation with diffusion model},
  author={Zhang, Mingyuan and Cai, Zhongang and Pan, Liang and Hong, Fangzhou and Guo, Xinying and Yang, Lei and Liu, Ziwei},
  journal={IEEE transactions on pattern analysis and machine intelligence},
  volume={46},
  number={6},
  pages={4115--4128},
  year={2024},
  publisher={IEEE}
}

@article{khosla2020supervised,
  title={Supervised contrastive learning},
  author={Khosla, Prannay and Teterwak, Piotr and Wang, Chen and Sarna, Aaron and Tian, Yonglong and Isola, Phillip and Maschinot, Aaron and Liu, Ce and Krishnan, Dilip},
  journal={Advances in neural information processing systems},
  volume={33},
  pages={18661--18673},
  year={2020}
}

@inproceedings{guo2022generating,
  title={Generating diverse and natural 3d human motions from text},
  author={Guo, Chuan and Zou, Shihao and Zuo, Xinxin and Wang, Sen and Ji, Wei and Li, Xingyu and Cheng, Li},
  booktitle={Proceedings of the IEEE/CVF conference on computer vision and pattern recognition},
  pages={5152--5161},
  year={2022}
}

@article{mason2022real,
  title={Real-time style modelling of human locomotion via feature-wise transformations and local motion phases},
  author={Mason, Ian and Starke, Sebastian and Komura, Taku},
  journal={Proceedings of the ACM on Computer Graphics and Interactive Techniques},
  volume={5},
  number={1},
  pages={1--18},
  year={2022},
  publisher={ACM New York, NY, USA}
}

@inproceedings{petrovich2022temos,
  title={Temos: Generating diverse human motions from textual descriptions},
  author={Petrovich, Mathis and Black, Michael J and Varol, G{\"u}l},
  booktitle={European Conference on Computer Vision},
  pages={480--497},
  year={2022},
  organization={Springer}
}

@inproceedings{athanasiou2022teach,
  title={Teach: Temporal action composition for 3d humans},
  author={Athanasiou, Nikos and Petrovich, Mathis and Black, Michael J and Varol, G{\"u}l},
  booktitle={2022 International Conference on 3D Vision (3DV)},
  pages={414--423},
  year={2022},
  organization={IEEE}
}

@inproceedings{chen2023executing,
  title={Executing your commands via motion diffusion in latent space},
  author={Chen, Xin and Jiang, Biao and Liu, Wen and Huang, Zilong and Fu, Bin and Chen, Tao and Yu, Gang},
  booktitle={Proceedings of the IEEE/CVF conference on computer vision and pattern recognition},
  pages={18000--18010},
  year={2023}
}

@incollection{hsu2005style,
  title={Style translation for human motion},
  author={Hsu, Eugene and Pulli, Kari and Popovi{\'c}, Jovan},
  booktitle={ACM SIGGRAPH 2005 Papers},
  pages={1082--1089},
  year={2005}
}

@inproceedings{min2010synthesis,
  title={Synthesis and editing of personalized stylistic human motion},
  author={Min, Jianyuan and Liu, Huajun and Chai, Jinxiang},
  booktitle={Proceedings of the 2010 ACM SIGGRAPH symposium on Interactive 3D Graphics and Games},
  pages={39--46},
  year={2010}
}

@article{holden2016deep,
  title={A deep learning framework for character motion synthesis and editing},
  author={Holden, Daniel and Saito, Jun and Komura, Taku},
  journal={ACM Transactions on Graphics (ToG)},
  volume={35},
  number={4},
  pages={1--11},
  year={2016},
  publisher={ACM New York, NY, USA}
}

@article{holden2017fast,
  title={Fast neural style transfer for motion data},
  author={Holden, Daniel and Habibie, Ikhsanul and Kusajima, Ikuo and Komura, Taku},
  journal={IEEE computer graphics and applications},
  volume={37},
  number={4},
  pages={42--49},
  year={2017},
  publisher={IEEE}
}

@article{aberman2020unpaired,
  title={Unpaired motion style transfer from video to animation},
  author={Aberman, Kfir and Weng, Yijia and Lischinski, Dani and Cohen-Or, Daniel and Chen, Baoquan},
  journal={ACM Transactions On Graphics (TOG)},
  volume={39},
  number={4},
  pages={64--1},
  year={2020},
  publisher={ACM New York, NY, USA}
}

@article{jang2022motion,
  title={Motion puzzle: Arbitrary motion style transfer by body part},
  author={Jang, Deok-Kyeong and Park, Soomin and Lee, Sung-Hee},
  journal={ACM Transactions on Graphics (TOG)},
  volume={41},
  number={3},
  pages={1--16},
  year={2022},
  publisher={ACM New York, NY}
}

@article{dhariwal2021diffusion,
  title={Diffusion models beat gans on image synthesis},
  author={Dhariwal, Prafulla and Nichol, Alexander},
  journal={Advances in neural information processing systems},
  volume={34},
  pages={8780--8794},
  year={2021}
}

@inproceedings{kim2023flame,
  title={Flame: Free-form language-based motion synthesis \& editing},
  author={Kim, Jihoon and Kim, Jiseob and Choi, Sungjoon},
  booktitle={Proceedings of the AAAI Conference on Artificial Intelligence},
  volume={37},
  number={7},
  pages={8255--8263},
  year={2023}
}

@article{ha2016hypernetworks,
  title={Hypernetworks},
  author={Ha, David and Dai, Andrew and Le, Quoc V},
  journal={arXiv preprint arXiv:1609.09106},
  year={2016}
}

@article{ho2020denoising,
  title={Denoising diffusion probabilistic models},
  author={Ho, Jonathan and Jain, Ajay and Abbeel, Pieter},
  journal={Advances in neural information processing systems},
  volume={33},
  pages={6840--6851},
  year={2020}
}

@article{song2020denoising,
  title={Denoising diffusion implicit models},
  author={Song, Jiaming and Meng, Chenlin and Ermon, Stefano},
  journal={arXiv preprint arXiv:2010.02502},
  year={2020}
}

@inproceedings{perez2018film,
  title={Film: Visual reasoning with a general conditioning layer},
  author={Perez, Ethan and Strub, Florian and De Vries, Harm and Dumoulin, Vincent and Courville, Aaron},
  booktitle={Proceedings of the AAAI conference on artificial intelligence},
  volume={32},
  number={1},
  year={2018}
}

@article{ho2022classifier,
  title={Classifier-free diffusion guidance},
  author={Ho, Jonathan and Salimans, Tim},
  journal={arXiv preprint arXiv:2207.12598},
  year={2022}
}

@article{jiang2023motiongpt,
  title={Motiongpt: Human motion as a foreign language},
  author={Jiang, Biao and Chen, Xin and Liu, Wen and Yu, Jingyi and Yu, Gang and Chen, Tao},
  journal={Advances in Neural Information Processing Systems},
  volume={36},
  pages={20067--20079},
  year={2023}
}

@article{ryu2023low,
  title={Low-rank adaptation for fast text-to-image diffusion fine-tuning},
  author={Ryu, Simo},
  journal={Low-rank adaptation for fast text-to-image diffusion fine-tuning},
  volume={3},
  year={2023}
}

@article{salimans2022progressive,
  title={Progressive distillation for fast sampling of diffusion models},
  author={Salimans, Tim and Ho, Jonathan},
  journal={arXiv preprint arXiv:2202.00512},
  year={2022}
}

@inproceedings{dai2024motionlcm,
  title={Motionlcm: Real-time controllable motion generation via latent consistency model},
  author={Dai, Wenxun and Chen, Ling-Hao and Wang, Jingbo and Liu, Jinpeng and Dai, Bo and Tang, Yansong},
  booktitle={European Conference on Computer Vision},
  pages={390--408},
  year={2024},
  organization={Springer}
}

@inproceedings{sampieri2024length,
  title={Length-aware motion synthesis via latent diffusion},
  author={Sampieri, Alessio and Palma, Alessio and Spinelli, Indro and Galasso, Fabio},
  booktitle={European Conference on Computer Vision},
  pages={107--124},
  year={2024},
  organization={Springer}
}

@misc{kingma2017adammethodstochasticoptimization,
      title={Adam: A Method for Stochastic Optimization}, 
      author={Diederik P. Kingma and Jimmy Ba},
      year={2017},
      eprint={1412.6980},
      archivePrefix={arXiv},
      primaryClass={cs.LG},
      url={https://arxiv.org/abs/1412.6980}, 
}

@article{elfwing2018silu,
  title={Sigmoid-weighted linear units for neural network function approximation in reinforcement learning},
  author={Elfwing, Stefan and Uchibe, Eiji and Doya, Kenji},
  journal={Neural networks},
  volume={107},
  pages={3--11},
  year={2018},
  publisher={Elsevier}
}

@inproceedings{raab2024monkey,
  title={Monkey see, monkey do: Harnessing self-attention in motion diffusion for zero-shot motion transfer},
  author={Raab, Sigal and Gat, Inbar and Sala, Nathan and Tevet, Guy and Shalev-Arkushin, Rotem and Fried, Ohad and Bermano, Amit Haim and Cohen-Or, Daniel},
  booktitle={SIGGRAPH Asia 2024 Conference Papers},
  pages={1--13},
  year={2024}
}

@inproceedings{guo2025stylemotif,
  title={StyleMotif: Multi-Modal Motion Stylization using Style-Content Cross Fusion},
  author={Guo, Ziyu and Lee, Young Yoon and Liu, Joseph and Ben-Shabat, Yizhak and Zordan, Victor and Kapadia, Mubbasir},
  booktitle={Proceedings of the IEEE/CVF International Conference on Computer Vision},
  pages={13349--13359},
  year={2025}
}

@article{zhong2025smoogpt,
  title={Smoogpt: Stylized motion generation using large language models},
  author={Zhong, Lei and Yang, Yi and Li, Changjian},
  journal={arXiv preprint arXiv:2509.04058},
  year={2025}
}

@inproceedings{zhang2024generative,
  title={Generative motion stylization of cross-structure characters within canonical motion space},
  author={Zhang, Jiaxu and Chen, Xin and Yu, Gang and Tu, Zhigang},
  booktitle={Proceedings of the 32nd ACM International Conference on Multimedia},
  pages={7018--7026},
  year={2024}
}

@article{guo2024generative,
  title={Generative human motion stylization in latent space},
  author={Guo, Chuan and Mu, Yuxuan and Zuo, Xinxin and Dai, Peng and Yan, Youliang and Lu, Juwei and Cheng, Li},
  journal={arXiv preprint arXiv:2401.13505},
  year={2024}
}

@inproceedings{karunratanakul2023guided,
  title={Guided motion diffusion for controllable human motion synthesis},
  author={Karunratanakul, Korrawe and Preechakul, Konpat and Suwajanakorn, Supasorn and Tang, Siyu},
  booktitle={Proceedings of the IEEE/CVF international conference on computer vision},
  pages={2151--2162},
  year={2023}
}

@article{xie2023omnicontrol,
  title={Omnicontrol: Control any joint at any time for human motion generation},
  author={Xie, Yiming and Jampani, Varun and Zhong, Lei and Sun, Deqing and Jiang, Huaizu},
  journal={arXiv preprint arXiv:2310.08580},
  year={2023}
}

@inproceedings{karunratanakul2024optimizing,
  title={Optimizing diffusion noise can serve as universal motion priors},
  author={Karunratanakul, Korrawe and Preechakul, Konpat and Aksan, Emre and Beeler, Thabo and Suwajanakorn, Supasorn and Tang, Siyu},
  booktitle={Proceedings of the IEEE/CVF Conference on Computer Vision and Pattern Recognition},
  pages={1334--1345},
  year={2024}
}

\end{document}